\documentclass[fleqn,10pt]{JLA_article}
\usepackage[english]{babel}
\usepackage{booktabs}
\usepackage{amsmath}
\usepackage{hyperref}
\hypersetup{hidelinks,colorlinks,breaklinks=true,urlcolor=color2,citecolor=color1,linkcolor=color1,bookmarksopen=false,pdftitle={LLM Judges as Raters},pdfauthor={Veerendra Kumar Sunkavalli}}

\PaperTitle{LLM Judges as Raters: A Pre-Registered Audit of Severity, Halo, Reliability, and Version Instability in LLM Essay Scoring on Public Corpora}
\Authors{Veerendra Kumar Sunkavalli}
\affiliation{\textit{Independent Researcher --- veeru.svk@gmail.com --- this work was conducted independently of the author's employment. Preprint; cite the peer-reviewed version once available.}}
\Submitted{28 August 2026}
\Accepted{---}
\Published{---}
\Volume{-}\Number{-}\Pages{1---14}\Doi{---}

\Notesname{Notes for Practice}
\note{LLM essay graders differ enormously in severity: two models that correlate equally well with human scores disagreed by up to 219 points (of 1000) about the same essays. Calibrate against your own human-anchored sample, on your own rubric wording (severity orderings were prompt-robust in Portuguese but not English), before any absolute use; 30--100 anchor essays recover most of the recoverable error. [confirmatory + tested remedy]}
\note{Hosted judges are mutable instruments (our evidence is from one serving platform; contrasts bundle weights with serving stack): a same-family version upgrade shifted scores by up to 13\% of the scale, and one model version became inaccessible to us mid-study when its provider legacy-gated it. Pin versions, re-score a fixed anchor set on a schedule (a paired 20-essay monitor caught 4 of our 5 shifts), and re-equate before mixing scores across versions. [confirmatory + tested remedy]}
\note{Averaging repeated calls makes a judge's score stable but not more accurate: replication sharpens a biased number. Budget replications for stability and human-anchored calibration for validity. [confirmatory]}
\note{Analytic sub-scores produced in a single call share a common impression (halo); whether it exceeds trained human raters is still open (our own matched comparison reversed under a robustness check), but scoring each dimension in a separate call halved it in our test. Validate diagnostic profiles per judge on your own instrument. [exploratory + tested remedy]}

\Abstract{Large language models (LLMs) are increasingly used as essay graders in learning analytics, evaluated mainly with agreement statistics. Educational measurement warns that raters also differ in severity, show halo, and drift. We treat LLM judges as raters and run a pre-registered rater-effects battery (severity, halo, generalizability/decision studies, version shifts, differential functioning) on public corpora in two languages (ENEM/Essay-BR; ASAP): 2,377 essays, 12 judges, 4 providers, 5 version contrasts, released as a score tensor. Judge severity spans 219 points on ENEM's 0--1000 scale; on ASAP the panel spread is 15--33\% of the score range (a max--min over 11 judges); matched on dispersion, judge severity SD is 8--15$\times$ that of trained raters. Judge--human correlations sit in a narrow .47--.56 band. All five version contrasts shift severity beyond a family-wise permutation null (up to 133 points), and one judge was legacy-gated mid-study, caught by identity canaries. Two pre-registered tests returned honest nulls: severity-adjusted leaderboard reversals did not survive a permutation null, and ``silent drift'' was refuted---agreement moved with severity in four of five contrasts. Replication yields self-consistency ($\phi\geq.80$ at $k\leq2$) but not human-level accuracy; a same-instrument check overturned our own halo comparison---no credible evidence that judge halo exceeds the trained-human range.}

\begin{document}
\flushbottom
\maketitle
\thispagestyle{empty}

\section{Introduction}
Learning analytics systems increasingly delegate the scoring of student writing to large language
models (LLMs). The dominant evaluation practice, inherited from automated essay scoring (AES),
asks one question: \emph{does the model agree with humans?}---operationalized as quadratically
weighted kappa (QWK) or a correlation against human scores. This question is necessary but
radically incomplete. A century of educational measurement on \emph{human} raters shows that
raters who correlate equally well with a criterion can still differ systematically in
\emph{severity} (how harshly they map quality onto the scale), exhibit \emph{halo} (letting a
holistic impression bleed into analytic sub-scores), and behave as \emph{unstable instruments}
over occasions \cite{saal1980,myford2003,myford2004,engelhard1994,eckes2015}. When scores carry absolute
consequences---placement, feedback, pass/fail thresholds, longitudinal dashboards---these rater
effects, not rank agreement, determine whether learners are treated fairly.

This paper treats LLM judges as what they operationally are in such pipelines: \emph{raters}, and
audits them with the classical rater-effects toolkit under a design that is impossible with human
raters: a fully crossed judges~$\times$~essays grid with replications, at a scale (over 100{,}000 scoring events) no human rater panel could deliver. Concretely, we score 977 ENEM
(Exame Nacional do Ensino M\'edio) essays from the public Essay-BR corpus \cite{marinho2021}
on the five official ENEM competencies, and 1,400 essays from four ASAP prompts
\cite{hewlett2012}, with 12 LLM judges spanning four providers, three price tiers, and five version contrasts (four within-family steps: sonnet 4$\to$4.5, 4.5$\to$4.6, opus 4.5$\to$4.6, nova-lite$\to$nova-2-lite; one same-provider cross-generation step: llama 3.3$\to$4), each cell replicated $K{=}3$ times (deep cell $K{=}5$). We analyze
the resulting tensor with many-facet Rasch measurement (MFRM) \cite{linacre1989,eckes2015},
residual halo analysis \cite{myford2003,myford2004}, generalizability and decision studies
\cite{brennan2001}, family-wise permutation tests for cross-version severity shifts, and a
purified cross-corpus differential-functioning analysis.

The study was pre-registered before any data collection (every one of the 110{,}571 captured calls is timestamped after the frozen protocol) (hypotheses H1--H6, decision rules,
exclusion rules, and analysis specifications frozen; all deviations logged and disclosed in
Appendix~\ref{app:dev}), and we report its two confirmatory nulls with the same prominence as its
positive results.

\textbf{Findings.} (1)~\emph{Severity is the dominant rater effect}: judge severity spans 219 points on the ENEM 0--1000 scale (11/12 judges displaced from the reference at a 99\% bootstrap criterion); the judge variance component is the second largest after essays (16.7\% of total); on ASAP the panel spread is 15--33\% of the score range where two trained humans differ by about 1\%. Ordering is robust to rubric paraphrase in Portuguese ($\tau{=}.89$) but not English ($\tau{=}.56$), itself a fragility finding. Meanwhile all judges sit in a narrow correlation band (.47--.56): location-blind statistics cannot see the largest effect we measure. (2)~\emph{Judges are unstable instruments}: all five version contrasts shift severity beyond a family-wise permutation null (up to 133/1000), and one judge was legacy-gated by its provider mid-study, caught by our canaries. (3)~\emph{Two pre-registered nulls, reported as such}: no severity-adjusted leaderboard reversal survives a permutation null (largest .029 vs.\ threshold .107), and ``silent drift'' was refuted: agreement moved with severity in four of five contrasts. (4)~\emph{Replication buys self-consistency, not validity} ($\phi{\geq}.80$ at $k{\leq}2$, largely temperature-injected variance; accuracy still loses to a second trained human on 3/4 ASAP instruments and on ENEM). A post-registration treatment arm (Section~\ref{sec:treat}) tests the remedies: anchored calibration removes the offset (the worst-case cutscore false-fail share 56\%$\to$10\%; a 133-point spurious cohort decline $\to$ $-0.1$) but not judge-specific disagreement; a paired 20-essay severity monitor detects four of five version shifts (power at least .86; the smallest shift needs about 100); per-dimension calls halve halo. (5)~\emph{Halo}: the pre-registered cross-instrument comparison suggested judges at or above the trained-human range; our same-instrument robustness check found no credible excess (no judge exceeds the more halo-prone rater on either matched instrument); a cautionary, still-open result reported in full.

\textbf{Contributions.} (i)~The first \emph{crossed, replicated} rater-effects audit of LLM essay
judges framed in the psychometric tradition (severity/fit, halo, G-theory, differential
functioning), on public education data in two languages; (ii)~a pre-registered design with
identity canaries, contamination audit with synthetic negative controls (0/22 cells flagged), and
two honestly-reported confirmatory nulls; (iii)~a released artifact: the full score tensor
(110,571 calls with verbatim raw captures), frozen rubrics in both languages, canary logs, and
analysis code, enabling reuse as a benchmark for judge calibration methods; (iv)~actionable
guidance for learning-analytics practice (Section~\ref{sec:practice}).

\section{Related Work}
\label{sec:related}
\textbf{Rater effects in educational measurement.} Severity/leniency, halo, central tendency, and
rater drift are classical, extensively documented effects in human essay rating
\cite{saal1980,engelhard1994,myford2003,myford2004,eckes2015}. MFRM \cite{linacre1989} models raters as a
facet, separating rater severity from examinee ability and task difficulty; generalizability
theory \cite{brennan2001} partitions score variance among raters, tasks, and occasions and
projects the reliability of alternative designs (D-studies). We import this toolkit wholesale.

\textbf{Writing analytics and automated writing evaluation in LA.} Automated evaluation of student writing has a long history in and around this venue, from operational AES systems \cite{attali2006,shermis2013,ramesh2022} to formative writing-analytics tools that foreground feedback over scores \cite{gibson2017,knight2020}. That literature has always carried a dual burden: technical agreement with human raters, and validity for the learning-support decisions built on top. LLM judges are now entering the same pipelines---often replacing purpose-built AES models on the strength of agreement statistics alone---which is precisely the gap this audit addresses: the properties that matter for feedback, placement, and progress monitoring are rater properties (severity, its non-uniformity, halo, stability), not rank agreement. Concretely: halo and compression distort the diagnostic profile formative feedback shows a student; severity and its origin determine who crosses a placement threshold; version instability masquerades as learning gains or losses on longitudinal dashboards---each rater effect in our battery maps onto a specific LA failure mode.\\[4pt]
\textbf{LLM-as-a-judge meta-evaluation.} A fast-growing literature examines LLM judges'
reliability and biases. Closest to us are five threads, from which we explicitly delineate:
(1)~psychometric \emph{datasheets/protocols} for judges \cite{judgedatasheet2026} propose what to
report but do not run a crossed rater-effects experiment; (2)~IRT analyses of LLM graders on
short-answer scoring \cite{irtasag2026} model item/grader ability but not rater facets, halo,
replication, or versions; (3)~agreement-metric guidance \cite{agreementmetrics2026} improves
reporting of exactly the statistics whose blind spots we quantify; (4)~G-theory has been applied to LLM autoscoring reliability in large-scale writing assessment
\cite{gtheoryap2025} and to LLM output nondeterminism in brand measurement \cite{brandvar2026},
establishing the variance-component framing; neither crosses a multi-provider judge panel with
essays and replications, nor joins it to severity/halo/version analyses;
(5)~MFRM used to correct \emph{human} labels in AI pipelines \cite{humanmfrm2026} is the mirror
image of our design (humans as raters; we model LLM judges as raters). Halo in LLM analytic
scoring has been flagged and addressed with alternative evaluation framings on ICNALE
\cite{selfref2026}; we instead \emph{quantify} halo via MFRM residual structure against
design-matched trained-human baselines. Descriptive studies of LLM--human grading differences
\cite{llmnothuman2026} observe severity-like biases informally; our contribution is their formal,
pre-registered psychometric quantification, with uncertainty and family-wise error control.

\begin{table*}[t]\centering
\caption{Positioning against the nearest prior work: what each contributes, and what this study adds.}
\label{tab:delta}\scriptsize
\begin{tabular}{p{4.6cm}p{5.2cm}p{5.6cm}}\toprule
Prior work & Contribution & Delta of this study \\ \midrule
Judge datasheet \cite{judgedatasheet2026} & Psychometric reporting protocol for LLM judges & A crossed rater-effects \emph{experiment} (severity/fit, halo, G/D-studies, versions), not a protocol \\
IRT on LLM short-answer graders \cite{irtasag2026} & Grader ability/difficulty via IRT (ASAG) & Rater \emph{facets} on essay scoring: severity CIs, halo, replications, version pairs, two languages \\
Agreement-metric guidance \cite{agreementmetrics2026} & What agreement statistics to report & Quantifies what those statistics cannot see (219-pt severity inside a .09-wide $r$ band) \\
G-theory for LLM autoscoring \cite{gtheoryap2025} & Variance components for LLM writing scores & Multi-provider crossed panel + severity/halo/version battery + trained-human benchmarks \\
Variance components of LLM nondeterminism \cite{brandvar2026} & D-study allocation for brand answers & Education measurement; judges as raters; validity (H5b) separated from stability (H5a) \\
MFRM correction of human labels \cite{humanmfrm2026} & Humans as raters in AI pipelines & The mirror design: LLMs as raters, human-anchored origin, cross-version instability \\
Halo in LLM L2 analytic scoring \cite{selfref2026} & Flags halo; proposes profile-based evaluation & Quantifies halo via matched residual statistic against design-matched trained-rater baselines \\
\bottomrule\end{tabular}\end{table*}

Table~\ref{tab:delta} makes the positioning explicit. We claim neither the first psychometric lens on LLM judges nor the first G-theory or halo
observation; the delta is the combination---a fully crossed judges$\times$essays$\times$replications
design, the complete classical battery, cross-lingual public education data, version pairs, and a
released tensor---plus pre-registration with honest nulls.

\section{Method}
\label{sec:method}
\medskip\noindent\textbf{Measurement concepts used in this paper (plain-language).} \emph{Severity}: a rater's systematic tendency to score low (or high) relative to a reference, here reported both in logits (model scale) and raw points. \emph{Halo}: correlation among a rater's sub-scores beyond what true ability structure explains, estimated from model residuals. \emph{Dependability} $\phi(k)$: how reproducible a score is if the same judge rescored $k$ times---consistency, not correctness. \emph{Family-wise permutation null}: a threshold set so that scanning many judge pairs cannot manufacture an effect.

\subsection{Corpora}

\textbf{Essay-BR (ENEM, Portuguese; primary).} The extended Essay-BR corpus \cite{marinho2021}
contains 6{,}577 student essays scored on ENEM's five competencies (C1--C5, each
$\{0,40,\dots,200\}$; total 0--1000). We sampled the 8 largest prompts $\times$ up to 125 essays,
stratified by human total after excluding essays under 50 words: 977 essays. Corpus scores come from public essay-correction platforms (student-submitted practice essays
corrected by the platforms' evaluators against the official five-competency rubric; the corpus
releases one score vector per essay with no rater identity or panel documentation; \citeNP{marinho2021}). We therefore treat the pooled human score as a \emph{reference
for the scale origin}, not as a rater with interpretable fit, and we verified that every
severity conclusion except the location of zero is origin-robust: re-expressing severity against
the pooled-judge centroid instead of the human reference preserves the ordering exactly
(Kendall $\tau = 1.0$) and the 219-point spread to one decimal. Only statements of the form
``judge X is more severe than \emph{the human standard}'' inherit the reference's limitations.
\textbf{ASAP (English; secondary).} From the Hewlett ASAP corpus \cite{hewlett2012} we use the
four genuine essay-writing sets (1, 2, 7, 8; the others are source-dependent short responses),
350 essays per set, stratified. ASAP provides \emph{two independent trained raters per essay},
which we use directly: as human severity/agreement baselines and (sets 7--8, trait scores) as a
design-matched human halo baseline. ASAP essay text cannot be redistributed under the original
competition rules; our artifact ships scores keyed by official essay IDs plus a hydration script
that verifies the source file by checksum. Essay-BR is MIT-licensed and shipped in full.

\subsection{Judges, replication, and instrumentation integrity}
\begin{table}[t]\centering\caption{Judge roster: 12 models, 4 providers, 3 price tiers, 5 version contrasts: 4 within-family steps + 1 same-provider cross-generation step (llama). Full platform identifiers, per-call response metadata, and canary logs are in the artifact.}
\label{tab:roster}\small\begin{tabular}{llll}\toprule
Judge & Provider & Tier & Version-pair role \\ \midrule
sonnet-4 & Anthropic & mid & sonnet steps; legacy-gated mid-study \\
sonnet-4.5 & Anthropic & mid & sonnet steps \\
sonnet-4.6 & Anthropic & mid & sonnet steps \\
opus-4.5 & Anthropic & premium & opus step \\
opus-4.6 & Anthropic & premium & opus step \\
haiku-4.5 & Anthropic & cheap & -- \\
nova-lite & Amazon & cheap & nova step \\
nova-2-lite & Amazon & cheap & nova step \\
nova-pro & Amazon & mid & -- \\
llama3.3-70b & Meta & cheap & cross-generation step \\
llama4-mav & Meta & cheap & cross-generation step \\
qwen3-32b & Qwen & cheap & -- \\
\bottomrule\end{tabular}\end{table}

Table~\ref{tab:roster} lists the panel: three Anthropic Claude Sonnet versions (4, 4.5, 4.6), two Opus versions (4.5, 4.6), Haiku 4.5, three Amazon Nova models (Lite, 2-Lite, Pro), two Meta Llama generations (3.3-70B, 4-Maverick), and Qwen3-32B---twelve judges on a single API platform (Amazon Bedrock), four providers, three price tiers, and five version contrasts: four within-family steps (\texttt{sonnet-4}$\to$\texttt{4.5}$\to$\texttt{4.6}; \texttt{opus-4.5}$\to$\texttt{4.6}; \texttt{nova-lite}$\to$\texttt{nova-2-lite}) plus the same-provider cross-generation step \texttt{llama3.3}$\to$\texttt{llama4}. Each judge scored every essay $K{=}3$ times (deep cell $K{=}5$ for the cheap tier on two ENEM prompts). Every request and response was captured
verbatim in an append-only store, including response metadata (request IDs, latency, service
headers), which constitutes the artifact of record since the API exposes no sampling seed.
Because several vendor model identifiers are mutable aliases rather than immutable pins, we ran
\emph{identity canaries}---a fixed set of 12 essays scored at temperature 0 by every judge at
collection start, at the day boundary, and at collection end (three sweeps across the collection window). The canaries earned their keep: one judge
(\texttt{sonnet-4}) was gated as ``legacy'' by its provider between our Portuguese and English
collection windows and failed 12/12 canary probes (availability loss---the endpoint refused requests---which canaries detect trivially; their designed purpose is the harder case of silent behavioral shift); its (complete, canary-clean) ENEM data are
retained, it has no English data, and no substitute was enrolled (version-pair members were
pre-registered as non-substitutable). Scoring used a fixed rubric prompt per language (the official five ENEM competencies in Portuguese; per-set holistic guides distilled from the official ASAP scoring materials in English), requesting compact JSON at temperature 0.7 (a deployment-typical default that also makes replication variance estimable; a temperature-0 anchor cell calibrates its contribution) with a 100-token cap. Parse rate was 100.0\% for every judge on both corpora (110{,}571 captured calls in total, of which 97{,}246 were pre-registered and 84{,}373 confirmatory (the odd call is a single \texttt{sonnet-4} ASAP response that completed 95\,s \emph{after} the legacy gate first fired---a propagation race, captured and excluded from analysis); a cell-by-cell call manifest in the artifact closes the accounting exactly; zero missing cells in the analyzed grids); off-scale values (e.g., a 100 on the ENEM 0--200 competency scale) were snapped to the nearest legal category under a frozen midpoint-down rule, with raw values preserved. Auxiliary cells (per-cell sizes: bridge 125 essays $\times$ 11 judges; paraphrase 250$\times$11 PT and 350$\times$11 EN; temperature-0 anchor: 125$\times$6$\times$3 pre-registered on the cheap tier, extended post-registration to all 11 judges): a \emph{holistic-rubric bridge} on ENEM
(the 11 post-gate judges, single 0--1000 score) to separate rubric structure from corpus effects; a
\emph{paraphrased rubric form} in each language (robustness of severity ordering); and a
\emph{temperature-0 anchor} cell (share of replication variance injected by sampling temperature).

\subsection{Pre-registration, analysis battery, and disclosure}
Hypotheses H1--H6, decision rules, exclusion rules, contamination decision rules, and analysis
specifications were frozen before full collection, and amended once---before any confirmatory
statistic was computed and before the English collection launched---in response to an independent
design review; all subsequent deviations are enumerated in Appendix~\ref{app:dev}. The battery:
\textbf{(A1)} MFRM (rating-scale model; facets essay, judge, competency, replication) with the
pooled human score fixing the origin; severity confidence intervals from a 1,000-resample
essay-cluster bootstrap on an independent implementation, with a frozen cross-implementation
ordering check (Kendall $\tau \geq .9$; observed $\tau = 1.0$ over the 11 non-reference judges, \texttt{sonnet-4.6} anchoring the MML side; the raw-mean check covers all 12).
\textbf{(A2)} A split-half severity-adjusted QWK leaderboard test with a permutation
max-statistic null over all 66 judge pairs.
\textbf{(A3)} Halo as mean off-diagonal correlation of standardized MFRM residuals across the
five competencies, compared against the same statistic computed for ASAP sets 7--8 trained human
raters (trait scores)---a matched-statistic, design-matched baseline.
\textbf{(A5)} Variance components for (essay:prompt)$\times$judge with replications nested in
cells (ANOVA expected-mean-squares estimators; the design is balanced and complete), with
per-judge D-studies ($\phi(k)$, self-consistency) and accuracy against humans (bias, RMSE) at
$k \in \{1,3\}$, benchmarked against ASAP's between-human-rater error.
\textbf{(A6)} Cross-version severity deltas against a family-wise sign-flip permutation null
($\alpha{=}.01$) plus an equivalence test (TOST, $|\Delta\text{QWK}|<.02$) for the ``silent
drift'' conjunction.
\textbf{(A7)} A contamination audit on both corpora (temperature-0 continuation probes, 150-word
prefixes, normalized longest-common-subsequence overlap) against synthetic negative-control
essays, with a frozen flagging rule. \textbf{(A4, exploratory)} Cross-corpus differential judge functioning with human-side equating and iterative purification, decomposed by the bridge cell.
Per COPE/JLA policy we disclose that data collection, analysis pipelines, and manuscript drafting
were performed with substantial LLM assistance under human direction and review; the LLM judges
under study are the object of measurement, and all analysis code and raw captures are released.

\begin{table}[t]\centering
\caption{Complete map of pre-registered hypotheses to outcomes. H2 (descriptive) and H4 (exploratory) were re-scoped in the pre-registration amendment itself (before any confirmatory statistic and before English collection), not post hoc.}
\label{tab:hmap}\footnotesize
\begin{tabular}{llp{4.1cm}}\toprule
ID & Status at freeze (v1.1) & Outcome \\ \midrule
H1 severity & confirmatory & \textbf{Supported} (\S\ref{sec:severity}) \\
H2 leaderboard (orig.) & demoted to descriptive at v1.1 & reported via H2b \\
H2b reversal test & confirmatory & \textbf{Null} (\S\ref{sec:h2b}) \\
H3 halo & confirmatory & knife-edge cross-instrument pass; \textbf{reversed} by same-instrument check (\S\ref{sec:halo}) \\
H4 cross-corpus DJF & exploratory at v1.1 & reported (\S\ref{sec:djf}) \\
H5a dependability & confirmatory & \textbf{Supported}, temp-conditioned (\S\ref{sec:gtheory}) \\
H5b accuracy & confirmatory & \textbf{Not supported} (\S\ref{sec:gtheory}) \\
H6 silent drift & confirmatory & \textbf{Refuted}; severity leg 5/5 (\S\ref{sec:versions}) \\
\bottomrule\end{tabular}\end{table}

\section{Results}
\label{sec:results}
Table~\ref{tab:hmap} maps every frozen hypothesis to its outcome; hypothesis IDs retain their pre-registration numbering (H2 and H4 were re-scoped in the amendment itself, which is why confirmatory reporting uses H2b and treats H4 as exploratory).

\subsection{Severity: large, judge-specific, and invisible to rank statistics (H1: supported)}
\label{sec:severity}
\begin{figure}[t]\centering\includegraphics[width=\linewidth]{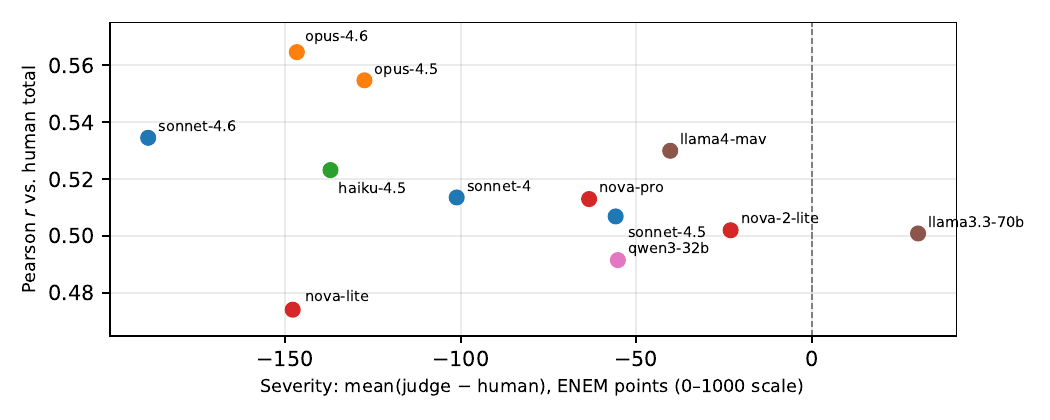}
\caption{Each point is one judge scoring the same 977 ENEM essays ($K{=}3$, rep-mean). Horizontal axis: severity (mean displacement from the human reference, points on the 0--1000 scale). Vertical axis: Pearson correlation with the human total. The largest rater effect in the study (219-point severity spread) is orthogonal to the statistic by which judges are commonly selected (all $r \in [.47,.56]$). Colors group model lines (sonnet, opus, haiku, nova, llama, qwen).}
\label{fig:severity}\end{figure}

\begin{table*}[t]\centering\caption{Judge-level rater effects on ENEM (977 essays, $K{=}3$). Severity point estimates and 99\% essay-cluster bootstrap CIs are both on the JMLE logit scale (human origin), so intervals and points are directly comparable (bootstrap distributions are left-skewed for the most severe judges, hence the asymmetric CIs); raw points on the 0--1000 scale; $r$ = Pearson correlation and QWK = quadratically weighted kappa vs. the human reference (QWK per competency on six categories, averaged); halo = mean off-diagonal residual inter-competency correlation; RMSE at $k{=}3$ (points/1000); $k^{*}$ = replications to reach $\phi \geq .80$ at temperature 0.7.}
\label{tab:severity}\small\begin{tabular}{lcccccccc}\toprule
Judge & Severity & 99\% CI & Raw pts & $r$ & QWK & Halo & RMSE$_{k=3}$ & $k^{*}$ \\ \midrule
llama3.3-70b & -0.38 & [-0.54, -0.18] & +30 & 0.50 & 0.38 & 0.38 & 183 & 1 \\
nova-2-lite & +0.28 & [0.11, 0.42] & -23 & 0.50 & 0.33 & 0.49 & 165 & 2 \\
llama4-mav & +0.49 & [0.30, 0.63] & -40 & 0.53 & 0.38 & 0.16 & 165 & 1 \\
qwen3-32b & +0.68 & [0.47, 0.82] & -55 & 0.49 & 0.24 & 0.38 & 166 & 1 \\
sonnet-4.5 & +0.71 & [0.53, 0.84] & -56 & 0.51 & 0.36 & 0.15 & 170 & 1 \\
nova-pro & +0.81 & [0.62, 0.93] & -63 & 0.51 & 0.31 & 0.38 & 167 & 2 \\
sonnet-4 & +1.28 & [1.06, 1.38] & -101 & 0.51 & 0.32 & 0.15 & 186 & 1 \\
opus-4.5 & +1.60 & [1.36, 1.66] & -127 & 0.55 & 0.36 & 0.22 & 204 & 1 \\
haiku-4.5 & +1.74 & [1.48, 1.80] & -137 & 0.52 & 0.32 & 0.19 & 210 & 1 \\
opus-4.6 & +1.83 & [1.58, 1.88] & -147 & 0.56 & 0.35 & 0.22 & 216 & 1 \\
nova-lite & +1.85 & [1.56, 1.91] & -148 & 0.47 & 0.18 & 0.25 & 216 & 2 \\
sonnet-4.6 & +2.34 & [2.03, 2.37] & -189 & 0.53 & 0.25 & 0.20 & 243 & 1 \\
\bottomrule\end{tabular}\end{table*}

Figure~\ref{fig:severity} and Table~\ref{tab:severity} summarize the central result. In raw score points---the primary reporting scale, since the measurement model is a descriptive projection (Appendix~A)---judge severity relative to the human origin spans 219 points on ENEM's 0--1000 scale: from \texttt{llama3.3-70b} at $+30$ (the only lenient judge) to \texttt{sonnet-4.6} at $-189$. On the JMLE/bootstrap logit scale the same effect spans 2.72 logits (SD $=0.76$, pre-registered criterion $>0.5$), and 11 of 12 judges exceed the frozen effect criterion ($|\hat\alpha_j| > 0.3$ logits with 99\% CI excluding 0.1), with
severity SD 64 points. The judge main effect is the second-largest variance component in the
G-study: 28\% as large as the pooled essay-plus-prompt component---29.5\% against within-prompt essay variance alone---and 16.7\% of total variance ($\hat\sigma^2_{\text{judge}} = 4{,}048$ vs.\ essay-plus-prompt 14{,}552, judge$\times$essay 3{,}471,
replication 1{,}877). \begin{table}[t]\centering\caption{Per-judge severity on ASAP, as percent of each set's score range (judge $k{=}3$ rep-mean minus the two trained raters' mean; negative = more severe than humans). Bottom row: the severity difference between the two trained human raters themselves --- the human yardstick. English severity orderings are prompt-conditional (paraphrase $\tau = .56$); the spread magnitude is more stable (set-1 spread 26\% of range under the paraphrased rubric vs. 33\%).}
\label{tab:asapsev}\small\begin{tabular}{lccccc}\toprule
Judge & Set 1 & Set 2 & Set 7 & Set 8 & Mean \\ \midrule
haiku-4.5 & -42.3 & -26.7 & -31.9 & -24.5 & -31.4 \\
nova-lite & -26.2 & -21.2 & -30.6 & -9.5 & -21.9 \\
sonnet-4.5 & -23.8 & -13.4 & -35.8 & -14.0 & -21.7 \\
llama4-mav & -25.2 & -12.2 & -38.9 & -8.2 & -21.1 \\
llama3.3-70b & -26.5 & -18.0 & -27.5 & -4.3 & -19.1 \\
sonnet-4.6 & -23.6 & -15.9 & -27.9 & -6.4 & -18.4 \\
opus-4.6 & -21.3 & -10.9 & -31.7 & -7.4 & -17.8 \\
opus-4.5 & -19.7 & -11.0 & -30.3 & -7.9 & -17.2 \\
nova-pro & -23.9 & -15.5 & -23.9 & -3.5 & -16.7 \\
qwen3-32b & -23.4 & -8.0 & -31.6 & -1.1 & -16.0 \\
nova-2-lite & -9.3 & -4.4 & -27.8 & -5.0 & -11.6 \\
\midrule
human rater 1 vs.\ rater 2 & -0.7 & -0.8 & -0.8 & -1.0 & -0.8 \\
\bottomrule\end{tabular}\end{table}

The one directly available human yardstick (Table~\ref{tab:asapsev}) puts the spread in perspective: on ASAP, the two trained raters of each set differ in mean severity by only 0.7--1.0\% of the score range, while the judge panel's severity spread on the same instruments is 15--33\% of range. These are extreme-order statistics (a max--min over 11 judges against a single two-rater gap), so we also report the matched comparison: the panel severity SD is 8--15$\times$ the SD implied by the trained pair, per set (per-set ratios 14.6/11.0/7.6/8.7 for sets 1/2/7/8: SD across the 11 judges' set means over the SD implied by the two trained raters, $|r_1-r_2|/\sqrt{2}$). The 219-point ENEM range itself is estimation-stable (essay-cluster bootstrap 95\% CI [212, 226]). Severity
\emph{ordering} is robust to an independently written Portuguese rubric paraphrase (Kendall $\tau = .89$, criterion $\geq .8$), so this is not a one-prompt artifact; the English paraphrase check \emph{failed} ($\tau = .56$), a scope condition flagged wherever English severity is discussed (and itself evidence of instrument fragility).

Against this 219-point spread, judge--human Pearson correlations occupy .47--.56 for all twelve judges, and QWK the band .18--.38 (computed per competency on the six rubric categories, then averaged; QWK is bin-sensitive, so the artifact fixes this rule). Location-invariant statistics ($r$, Spearman, rank leaderboards) are blind to severity \emph{by construction}; QWK is partially sensitive (offsets depress it through marginal mismatch), consistent with the QWK movement under version shifts in Section~4.6. Neither \emph{reports} severity: $r{=}.52$ or QWK $=.30$ says nothing about whether the judge sits 30 points above or 190 below the human scale.
The absolute level of the band is also attenuated by reference unreliability: on ASAP, where a
second trained rater calibrates the ceiling, the best judges reach $r = .51$--$.69$ against one
rater where the other human reaches $.62$--$.77$---so against well-anchored references the best
judges approach---and on one instrument exceed (set 8: .69 vs.\ .62)---human--human rank agreement, and the ENEM band should not be
read as an absolute performance ceiling. Location-invariant statistics are blind to severity by construction (an
exploratory but arithmetically inevitable observation): two judges indistinguishable by $r$ can disagree by more than a full rubric category per competency about every essay (219/5 competencies $=$ 44 points against a 40-point step). Contamination probes flagged 0/22
judge$\times$corpus cells against synthetic-control baselines, and MFRM severity orderings were
identical across two independent implementations ($\tau=1.0$).

\medskip\noindent\textbf{Severity is not a constant offset: differential severity by
essay quality (exploratory, RTM-checked).} Conditioning each judge's displacement on quality
bands suggests strong \emph{scale compression}: on ENEM, judges average $+62$ points above the
human score on the weakest quintile and $-216$ below on the strongest. Conditioning on an
error-laden reference, however, produces exactly this pattern by regression to the mean, so we
ran an RTM-robust check on ASAP, where two independent raters exist: essays are binned by
rater~1 and judge displacement is measured against rater~2 (whose error is independent of the
binning), with the rater-1-vs-rater-2 slope as a placebo. Compression survives decisively on
two of four instruments (set~1: 11/11 judges negative slopes, median $-0.72$ of a 5-point
range; set~7: 11/11, median $-3.28$ of 12---while the placebo slopes are \emph{positive}), is
absent on set~2, and \emph{reverses} on the long narrative set~8 (median $+2.04$ of 25).
Differential severity by quality is therefore real but instrument-dependent, and the ENEM
magnitudes should be read as RTM-inflated upper bounds. The practical point survives in
weakened but still consequential form: on some instruments a single additive calibration
cannot fix an offset that varies across the quality range (conditional, e.g.\
equipercentile, equating is required), and where compression holds it flatters the weakest
writers and penalizes the strongest---an equity-relevant pattern that deployments should test
for \emph{on their own instrument} rather than assume present or absent.

\subsection{Does severity overturn leaderboards? A pre-registered null (H2b: not supported)}
\label{sec:h2b}
Does severity of this magnitude corrupt judge \emph{selection}? Our pre-registered test says: not demonstrably, at this scale. Estimating per-judge, per-competency severity offsets on one random half of essays, subtracting them from the other half's scores, re-snapping to legal categories, and evaluating both the raw and adjusted QWK leaderboards on that held-out half produced 16 nominal pairwise rank reversals, but the largest reversal magnitude
(.029) falls far short of the permutation max-statistic null over the 66-pair family
(95\% threshold .107). Two honest qualifications: the design is deliberately conservative
(split-half estimation plus family-wise correction), so this null bounds---it does not exclude---small reversal effects (the family-wise threshold implies the design could only detect reversal magnitudes exceeding $\approx$.11 QWK); and adjacent QWK leaderboard ranks are noise-dominated in this
regime, so leaderboard positions themselves should carry uncertainty intervals, which current
practice rarely reports. Concretely, the raw QWK leaderboard on the evaluation half (competency-averaged QWK computed on that half only) is led by the two cheap Meta models (note: half-sample QWK values differ from the full-grid QWK column of Table~\ref{tab:severity}) (llama3.3-70b .367, llama4-mav .359), followed by opus-4.5 (.350), opus-4.6 (.338), and sonnet-4.5 (.336)---so even cheap models can lead; severity adjustment moves several adjacent pairs by one rank, but every such swap is smaller than what label-permuted severity corrections produce by chance at the family level. The artifact releases the full leaderboards and permutation distribution for better-powered designs to revisit. The practical conclusion is asymmetric: severity demonstrably matters
for \emph{absolute} uses of scores (Sections~\ref{sec:severity},~\ref{sec:gtheory}); we could not
demonstrate that it matters for \emph{comparative} judge rankings.

\subsection{Halo: the same-instrument check reverses the cross-instrument comparison (H3)}
\label{sec:halo}
\emph{Readers should treat the same-instrument robustness check at the end of this subsection as the estimate of record; the pre-registered cross-instrument comparison that follows is reported for completeness and is superseded by it.} On the matched residual statistic, judges' mean residual halo is $r = .264$, with large per-judge heterogeneity (range .15--.49; four judges above the high-halo human value of .26, eight below it). The design-matched human baseline---ASAP sets 7--8 trained raters, same JMLE-residual
computation on their trait scores---is heterogeneous: set-7 raters show near-zero residual halo
(.00--.07) while set-8 raters show .26. Judges therefore clearly exceed the low-halo human regime
and are statistically indistinguishable from the high-halo regime; against the (pre-registered as
pooled) baseline of .148, the judge mean clears the frozen $>.10$ margin by only .016 (per-judge 99\% CIs in the artifact), and 5/12 judges individually meet the criterion. Because the pass rule was not pre-specified and the pooled baseline averages two visibly different regimes, we report the cross-instrument comparison as a superseded estimate rather than a finding---with the added scope condition that the judge statistic comes from ENEM's five Portuguese competencies and the human statistic from ASAP's English trait rubrics, so residualization may remove true structure unequally across instruments. The ENEM pooled human scores' residual halo (.48) and all raw-scale halo values (judges .48--.79 rep-mean; pooled ENEM human .64; ASAP raters' raw trait halo .54--.74 on the sampled essays) are reported descriptively in the artifact; raw-scale values are dominated by true multidimensional ability correlation and are not evidence of rater halo on their own.

\medskip\noindent\textbf{Same-instrument robustness check (post-registration, exploratory).} The comparison above is cross-instrument by construction (ENEM competencies vs.\ ASAP traits), the one gap in an otherwise crossed design. We therefore collected judge \emph{trait} scores on ASAP sets 7--8 under the same trait structure the human raters used (7,700 additional calls, 100\% parse), and computed residual halo for judges and both human raters \emph{within a single calibration per instrument}. The direction reverses: judge halo is $.07$--$.36$ (median .19) on set 7 and $.27$--$.58$ (median .37) on set 8, against human raters at $.36/.47$ and $.53/.59$ respectively---no judge exceeds the more halo-prone rater on either instrument, and on set 7 none exceeds even the less halo-prone one (on set 8, two judges---nova-pro and qwen3-32b---sit between the two raters; full per-judge values in the artifact). Two lessons follow. Substantively: no credible evidence that judges exceed trained humans in halo once instrument and calibration are matched---though all such point comparisons, including this one, are calibration-sensitive, and we regard H3's substantive question as \emph{unresolved pending a registered same-instrument replication} rather than answered in the reversed direction. We also acknowledge a check-selection asymmetry: this robustness check was commissioned after the confirmatory direction was known, in response to an independent review of the instrument confound; a clean H3 failure might not have prompted it. Methodologically: residual-halo values are strongly calibration- and instrument-dependent (the same human raters show .00--.26 in a humans-only calibration and .36--.59 in the joint one), so halo comparisons across studies differing in either are uninterpretable---including our own pre-registered one.

\subsection{Reliability: self-consistency is cheap, validity is not (H5a: supported; H5b: not supported)}
\label{sec:gtheory}
\begin{table}[t]\centering\caption{Variance components, ENEM total score (ANOVA-EMS estimators; (essay:prompt)$\times$judge design with replications nested in cells; 977 essays $\times$ 12 judges $\times$ 3 replications, complete).}
\label{tab:gtheory}\small\begin{tabular}{lrr}\toprule
Component & $\hat\sigma^2$ & \% of total \\ \midrule
Prompt & 836 & 3.5\% \\
Essay (within prompt) & 13,716 & 56.7\% \\
Judge (severity) & 4,048 & 16.7\% \\
Judge $\times$ prompt & 245 & 1.0\% \\
Judge $\times$ essay & 3,471 & 14.3\% \\
Replication (within cell) & 1,877 & 7.8\% \\
\bottomrule\end{tabular}\end{table}

Table~\ref{tab:gtheory} reports the variance decomposition (note that essays were sampled stratified on the human total and the prompt facet rests on 8 prompts, so the essay component---and hence all ratios to it---are conditional on this design; unstratified sampling would plausibly enlarge the essay component and shrink the judge share). D-studies from the variance components give $\phi(k) \geq .80$ at $k{=}1$ for nine judges and $k{=}2$ for the remaining three; at temperature 0 the cheap tier reaches $\phi(1) = .90$--$1.00$ with no
replication at all: \emph{replication was never the bottleneck}. Moreover the temperature-0
anchor cell (extended post-registration to all 11 post-gate judges; 44--61\% for the mid/premium tier) shows 38--100\% of replication variance is injected by the sampling temperature we
chose (one judge is fully deterministic at $t{=}0$), so ``self-consistency'' is substantially a property of a decoding knob, and all $\phi(k)$ values are scope-conditioned on $t{=}0.7$; severity itself, by contrast, is temperature-stable (mean shift between $t{=}0$ and $t{=}0.7$ within $\pm 9$ points of 1{,}000 for all 11 judges).
The pre-registered accuracy claim is evaluated against each ASAP set's design-matched benchmark: the RMSE between its own two trained raters. The best judge beats that benchmark on one of four instruments (set 8: \texttt{qwen3-32b}, 2.63 vs.\ 2.74) and loses on the other three (set 1: 0.85 vs.\ 0.63; set 2: 0.69 vs.\ 0.52; set 7: 3.54 vs.\ 1.88; per-judge table in the artifact); since
the pre-registration did not fix the instrument quantifier, we disclose both readings and score
H5b \textbf{not supported} on the strict (majority-of-instruments) reading that we consider
the honest one. On ENEM no second human rating exists; the corresponding comparison
(range-normalized RMSE .165 vs.\ an ASAP-imported .131) is reported as heuristic only, and
removing each judge's mean severity bias narrows but does not close it (.153--.180,
exploratory)---the deficit is partly attributable to, but not eliminated by removing,
severity. In sum: LLM judges are \emph{precise} about scores that are \emph{wrong in a judge-specific direction}; replication sharpens a biased estimate.

\subsection{From diagnosis to treatment: calibration, cutscores, and monitoring (post-registration, exploratory)}
\label{sec:treat}
\begin{table}[t]\centering\caption{Treatment arm (post-registration, exploratory): panel-mean held-out RMSE (ENEM points; 200 splits, 400 eval essays, anchors $n{=}100$; 95\% split intervals in the artifact) and false-fail/false-pass shares (of all essays) for the most severe cheap judge (nova-lite) at the 600-point cutscore. Reference = pooled platform score (scale reference, not error-free ground truth).}
\label{tab:treat}\small\begin{tabular}{lcc}\toprule
Calibration & held-out RMSE & FF / FP (nova-lite) \\ \midrule
none (raw) & 190.7 & .56 / .01 \\
mean offset & 160.4 & .10 / .16 \\
linear equating & 154.2 & .10 / .16 \\
equipercentile & 177.5 & .14 / .14 \\
\bottomrule\end{tabular}\end{table}

A diagnosis of severity is only as useful as the remedy it licenses, so we tested our own advice on held-out data (the existing tensor, except the per-dimension halo cell, which required 3,750 new calls; none pre-registered; full specification, per-judge results, and CIs in Appendix~A and the artifact). \textbf{Calibration:} anchoring each judge on $n$ human-scored essays reduces panel-mean held-out RMSE from 191 (no calibration; mean per-judge 95\% split interval [181, 201]) to 167/162/160 with a mean offset at $n{=}10/30/100$ ([148, 173] at $n{=}100$), and to 154 with linear equating ([142, 167]); the offset-vs-linear difference is not separable at these anchor sizes, and equipercentile mapping overfits (178). Returns saturate quickly, but no method approaches zero: calibration removes the offset, not the judge-specific disagreement (both numbers anchored to the ENEM platform reference, which is a scale reference rather than error-free ground truth---the caveat of Section~3.1 applies to every absolute value here). \textbf{Cutscore consequences:} at a 600-point threshold on ENEM, under the most severe cheap judge (\texttt{nova-lite}), essays that pass by the reference yet fail under the judge amount to 56\% of all essays out of the box; offset calibration with 100 anchors cuts false-fails to 10\% (at 16\% false-passes)---a large improvement that still leaves 26\% of all essays misclassified at the threshold (false-fail .10 $+$ false-pass .16): even calibrated, no judge in this panel is fit for unassisted absolute cutscore decisions. Replicated with error-aware ground truth on ASAP set 1 (rater 2 as criterion, cut at 4 of 6): uncalibrated judges' pass-by-criterion-yet-failed shares span 11--86\% of all essays (median 66\%) where rater 1---a trained human under the same criterion---sits at 4\%. \textbf{Version boundaries:} a cohort re-scored across the sonnet-4.5$\to$4.6 step (Section~\ref{sec:versions}) with no true change would show a spurious 133-point (0.74 human-SD) decline; per-version anchoring on 100 essays removes it in expectation ($-0.1$ points), with a single-deployment calibration error of SD 6.6 points (95\% of deployments within $\pm$13). \textbf{Monitoring:} re-scoring the \emph{same} $m$ anchor essays under both versions and applying a paired $t$-test (empirical false-alarm .04--.06 at nominal $\alpha{=}.05$) detects four of the five version shifts with power $\geq.86$ at $m{=}20$ (empirical per-pair paired-delta SDs 47--104 points; full power grid in the artifact); the smallest shift (opus, $-19$ points) needs $m{\approx}100$ (power .99)---size the monitor to the smallest shift that matters for your decisions. A QWK monitor at matched false-alarm detects the large sonnet shift with power .77 at $m{=}20$ and .93 at $m{=}50$: agreement monitoring is workable but less efficient, and unlike the severity monitor it reports neither direction nor magnitude, which recalibration needs. \textbf{Halo treatment:} scoring each competency in a separate call (the previously untested advice of Note 4) roughly halves residual halo on the same essays (median .31$\to$.16 across the cheap tier; 5 of 6 judges reduced)---the advice survives its test.

\subsection{Version updates are loud instrument changes (H6 as pre-registered: refuted; severity component: 5/5)}
\label{sec:versions}
\begin{figure}[t]\centering\includegraphics[width=\linewidth]{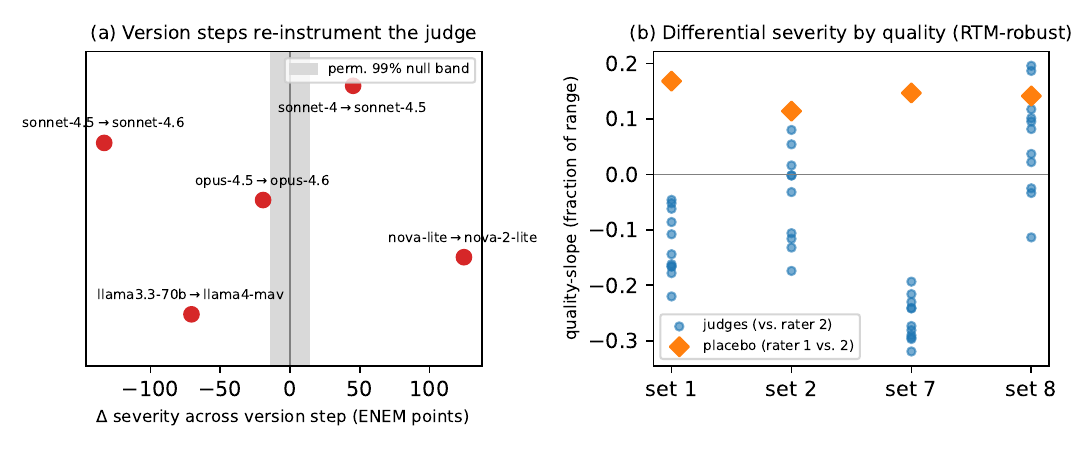}
\caption{(a) Severity shifts across the five version contrasts against the family-wise sign-flip permutation 99\% null band (ENEM, all pair members canary-clean). (b) RTM-robust differential severity: per-judge quality slopes (displacement vs.\ rater 2, essays binned by rater 1) as a fraction of each ASAP set's range; orange diamonds are the rater-1-vs-rater-2 placebo slopes, which have the opposite sign where compression holds.}
\label{fig:versions}\end{figure}
\begin{table}[t]\centering\caption{Cross-version severity shifts (ENEM). $\Delta$sev in raw points (positive = newer version more lenient); >null = exceeds the family-wise sign-flip permutation 99\% max-statistic; $\Delta$QWK 90\% bootstrap CI and TOST equivalence at $\pm.02$. The pre-registered silent-drift conjunction failed for every pair.}
\label{tab:versions}\small\begin{tabular}{lcccc}\toprule
Pair & $\Delta$sev & $>$null & $\Delta$QWK CI & equiv. \\ \midrule
sonnet-4 $\to$ sonnet-4.5 & +45 & yes & [+0.025, +0.059] & no \\
sonnet-4.5 $\to$ sonnet-4.6 & -133 & yes & [-0.128, -0.090] & no \\
opus-4.5 $\to$ opus-4.6 & -19 & yes & [-0.021, -0.001] & no \\
nova-lite $\to$ nova-2-lite & +125 & yes & [+0.122, +0.172] & no \\
llama3.3-70b $\to$ llama4-mav & -71 & yes & [-0.013, +0.030] & no \\
\bottomrule\end{tabular}\end{table}

The pre-registered ``silent drift'' hypothesis---that a version update shifts severity while QWK
stays equivalent---\emph{failed, informatively}: no contrast passed the TOST equivalence bound, and in four of the
five contrasts the 90\% bootstrap CI for $\Delta$QWK itself excludes zero (sonnet 4$\to$4.5:
$[+.025,+.059]$; 4.5$\to$4.6: $[-.128,-.090]$; opus: $[-.021,-.001]$; nova: $[+.122,+.172]$;
llama: $[-.013,+.030]$, inconclusive)---agreement demonstrably moved, not merely failed to prove
equivalent. What the data show instead is stronger and simpler: version updates
are \emph{loud} instrument changes. All five severity contrasts exceed the family-wise permutation null at $\alpha{=}.01$ (all four within-family steps, and the cross-generation llama step) (all estimated on the ENEM grid, which is complete and
canary-clean for every pair member including the subsequently-gated \texttt{sonnet-4};
excluding that pair leaves 4/4): \texttt{sonnet-4.5}$\to$\texttt{4.6} $=-133$ points
(a shift of 0.67 of a rubric-category step on each of the five competencies),
\texttt{nova-lite}$\to$\texttt{nova-2-lite} $=+125$, \texttt{llama3.3}$\to$\texttt{llama4}
$=-71$, \texttt{sonnet-4}$\to$\texttt{4.5} $=+45$, \texttt{opus-4.5}$\to$\texttt{4.6} $=-19$ (deltas computed from unrounded severities; recomputing from the rounded per-judge table values can differ by $\pm 1$ point).
Sign and magnitude vary within a provider and even within a family across steps: updates do not drift in a house direction; they re-instrument. Because all judges are served through one hosted platform with mutable aliases, each contrast is strictly a claim about \emph{hosted endpoints}---weights and serving stack jointly; disentangling weights from serving-stack changes requires pinned open-weights checkpoints (future work). Adjacent to this (an $n{=}1$ case observation,
not part of the hypothesis test): during the study the provider legacy-gated \texttt{sonnet-4} for accounts like ours---an audited example of the mutability of hosted judges, detected within hours by temperature-0 canaries whose logs we release.

\subsection{Contamination audit and quality control}
Both corpora are public, ASAP since 2012, so pre-training contamination could inflate agreement or distort rater effects. Under the frozen decision rule (temperature-0 continuation probes on 150-word prefixes of 50 essays per corpus for every judge, normalized longest-common-subsequence overlap, thresholded at 3 standard deviations above matched synthetic-control essays generated after every judge's training window), probed essays required $\geq$220 words so that a 150-word prefix leaves a scoreable continuation (nominal probes = (50 real + 50 synthetic) essays $\times$ 2 corpora $\times$ 11 judges = 2,200, one probe per essay$\times$judge; 1,936 passed the length filter, plus 100 control generations = 2,036 captured probe-cell calls); 0 of 22 probed judge$\times$corpus cells were flagged and neither corpus met the demotion criterion (22 of the 23 data-bearing cells---12 ENEM + 11 ASAP: \texttt{sonnet-4} was legacy-gated before the audit ran, so its retained ENEM results, including the sonnet-4$\to$4.5 severity delta, carry an unaudited-contamination caveat its 11 probed peers do not). We treat this as absence of \emph{detectable verbatim} memorization, not proof of no contamination: continuation probes are weak detectors, which is one reason the design weights the (younger, Portuguese) ENEM corpus as primary. Quality control: 100\% JSON parse rate for all judges on both corpora; identical severity orderings across two independent MFRM implementations ($\tau=1.0$); replication counts complete in every analyzed cell; per-judge category-usage histograms, retry counts, and response metadata are released.

\subsection{Cross-corpus differential judge functioning (exploratory)}
\label{sec:djf}
Severity is not a stable trait of a judge across corpora (per-judge DJF and bridge decompositions are tabled in the artifact). Expressed in human-SD units and equated
on the human side with iterative purification, judge severity moves between ENEM and ASAP by
$-0.84$ to $+0.42$ SD (5/11 judges flagged as differentially functioning). The holistic-rubric
bridge cell on ENEM shows that rubric structure alone (analytic vs.\ holistic, same essays, same
language) moves severity by $-0.79$ to $+0.41$ SD depending on judge---for several judges as large as the flagged cross-corpus effects themselves. Language, rubric structure,
scale, population, genre, human-rater regime, and residual contamination differences are
confounded in the corpus contrast (we enumerate them; the bridge isolates only the rubric
component), so these are exploratory magnitudes---but they caution against transporting a
severity calibration across contexts.

\section{Discussion and Implications for Learning Analytics}
\label{sec:practice}
\textbf{For researchers.} Report rank/agreement statistics alongside, not instead of, severity and its decision-scale consequences. The battery costs $\approx$\$460 of API calls and is fully scripted in the artifact. Our two nulls are as informative as our positives---claims about LLM-judge pathologies should be tested against family-wise nulls before entering the folklore.

\textbf{What the crossed design buys.} Every analysis in this paper is enabled by a design property human rater studies cannot have: the same 977 essays scored by all twelve raters three times each, with zero missing cells. With human raters, severity, halo, and reliability must be disentangled from massive missingness by the measurement model itself; here the model is a convenience, not a rescue, and every model-based claim can be checked against simple cell means (our severity orderings agree across an MML estimator, a JMLE estimator, and raw mean differences, $\tau = 1.0$). The marginal cost of this completeness was about \$460 in API calls: rater-effects audits can be a standard pre-deployment step, not a research luxury.\\[4pt]
\textbf{Why the nulls matter.} Both nulls constrain real design decisions. The H2b null says: if you are choosing \emph{which} judge to deploy by agreement metrics, we could detect no reversal above the design's $\approx$.11-QWK detection floor---so at practically consequential magnitudes, selection and calibration behave as separable problems. The H6 refutation says: you cannot rely on agreement metrics staying flat to detect version change (they move), but neither do you need to---the severity shift itself is directly measurable with a small anchored sample. We submit both nulls as family-wise-corrected data points against two plausible-sounding claims we ourselves pre-registered.\\[4pt]
\textbf{Notes for practitioners.} (1)~\emph{Calibrate the origin, not just the ranking:} before
using an LLM grader for absolute decisions, estimate its severity against a human-anchored sample
from \emph{your} population and rubric; expect per-judge offsets of 23--189 points of 1{,}000 (0.12--0.95 of a rubric category per competency), spanning 219 points across our panel, and do
not assume an offset measured on one rubric or language transports to another.
(2)~\emph{Pin and canary:} treat a hosted judge as a mutable instrument. Pin datestamped versions
where offered; run a fixed canary set at temperature 0 on a schedule; treat any distribution
shift as a re-calibration event. Expect versions to be legacy-gated or retired---one of ours became inaccessible to us mid-study.
(3)~\emph{Version updates are re-instrumentation events:} a model upgrade of the same family
shifted scores by up to 13\% of the ENEM scale in our data. Never mix scores across judge
versions in longitudinal analytics without re-equating.
(4)~\emph{Replication is not a validity fix:} averaging $k$ calls stabilizes the number without
moving it toward the human reference; budget replications for stability, and calibration for
truth. (5)~\emph{Analytic sub-scores from a single call share a common impression:} where diagnostic profiles matter, score dimensions in separate calls---this halved residual halo in our test (Section~\ref{sec:treat})---and validate that profile differences, not just totals, are trustworthy.

\textbf{Limitations.} Human references differ across corpora (pooled platform scores for ENEM;
double-blind trained raters for ASAP), and the ENEM origin is a population reference, not a rater
panel; severity estimates are conditional on one (paraphrase-validated, in PT) rubric prompt, one platform, one decoding regime (the 100-token cap forecloses rationale-first scoring, whose rater effects may differ), and a collection window; neither corpus carries demographics, so learner-subgroup DIF---the fairness analysis LA ultimately needs---is out of reach here; the English paraphrase check failed,
so English severity orderings should be treated as prompt-conditional; halo baselines span two
visibly different human regimes; H2b's null is bounded by a deliberately conservative test;
contamination probes (continuation-based, with synthetic controls) are a weak detector and their
cleanliness is not proof of no memorization; the un-datestamped platform aliases mean
within-version identity is certified only by canary windows; and all findings concern essay
scoring in two educational corpora, not LLM-judging in general. Transport beyond this platform is untested: pinned open-weights checkpoints would remove the mutable-alias risk, and serving-stack differences (quantization, batching) could shift severity by amounts our design cannot separate from the model itself; the audit recipe transfers unchanged.

\section{Conclusion}
Treated as raters, LLM essay judges show the classical pathology profile of untrained human raters (large idiosyncratic severity; halo with no credible excess over trained raters), compounded by a
failure mode humans do not have: unannounced, provider-controlled re-instrumentation, up to and
including mid-study disappearance. None of this is reported by the agreement statistics by which judges are currently selected, and none of it is fatal: severity can be measured, anchored, and monitored with decades-old measurement tools, at a cost of a few hundred dollars. The released tensor and harness are offered as a standing benchmark: any proposed judge,
prompt, or calibration method can be dropped into the same crossed design and audited against the
same battery. The paper succeeds if ``we used an LLM judge'' comes to be routinely followed by ``and here is its severity, measured this week.''

\section*{Ethics Statement}
This study uses only pre-existing public corpora of student writing and creates no new data about identifiable learners. Essay-BR essays were submitted by students to public essay-correction platforms and released by the corpus authors under an MIT license \cite{marinho2021}; ASAP essays were released for public research use in anonymized form in 2012, with named entities replaced by placeholders by the organizers \cite{hewlett2012}, and the original terms prohibit redistribution of essay text and any attempt to identify writers, both of which we honor (our artifact ships scores and identifiers, never ASAP text, and our contamination probes test verbatim memorization without any authorship inference). Released raw captures contain model-generated scores/JSON only; contamination-audit \emph{continuations} are not released (only overlap statistics), since they could partially reproduce non-redistributable text. Sending essay text to commercial scoring APIs mirrors the audited deployment practice; the Brazilian essays were already public with no personal identifiers, and all analyses characterize the \emph{judges}, never individual writers. The study used no human participants and required no ethics-board review under our institution's policy for public secondary data.

\section*{Data Availability}
Artifact (score tensor with verbatim raw captures, rubrics in both languages and paraphrase forms, manifests, canary logs, probe outputs, deviation log, and analysis code with pinned environments; an OSF mirror of the pre-registration, its amendment, and the deviation log is at \href{https://osf.io/zqhmf/?view_only=c18b761d72d247a2a92c62a7eba653ed}{\texttt{osf.io/zqhmf}} (view-only, key \texttt{c18b761d72d247a2\-a92c62a7eba653ed}; files uploaded 2026-08-28, after collection. The protocol-before-data ordering rests on the freeze record (sha256 \texttt{cb8e3775}\ldots\ in the deviation log) and on every captured call being timestamped after it; two limits: the freeze record lives in author-controlled version control (timestamp not independently attested), and the OSF hash match confirms byte-identity with the frozen protocol, not when it was frozen. A formal OSF Registration will be linked in revision):
\url{https://anonymous.4open.science/r/lak-27-A618} (view-only). Essay-BR text is included (MIT); ASAP text is restored by a checksum-verified hydration script.

\appendix
\renewcommand{\thesection}{\Alph{section}}
\titleformat{\section}{\color{color1}\large\bfseries}{}{0em}{\colorbox{color2!10}{\parbox{\dimexpr\linewidth-2\fboxsep\relax}{Appendix \Alph{section}. #1}}}
\section{Measurement Model and Instrumentation Details}
\label{app:model}
\textbf{MFRM specification.} For essay $n$, judge $j$, competency $d$, replication $r$: $\log(P_{ndjrk}/P_{ndjr(k-1)}) = \theta_n - \alpha_j - \delta_d - \rho_r - \tau_k$, with judge, competency, and replication facets summing to zero and the pooled human score fixing the origin ($\alpha_{\text{human}} \equiv 0$; the judges-sum-to-zero alternative was run as a pre-registered sensitivity and preserves every ordering and H1 verdict). Point estimates of record come from marginal maximum likelihood (TAM 4.3-25, R 4.6.1); confidence intervals come from a 1{,}000-resample essay-cluster bootstrap on an independent vectorized JMLE implementation whose severity orderings must agree with the MML estimates at Kendall $\tau \geq .9$ (observed: $\tau = 1.0$; the JMLE logit scale is less compressed than the MML scale, which is why raw-point severities are reported alongside everywhere). Fit statistics use the first replication only (pooled replications violate local independence); infit .11--.78 indicates strong local dependence---the phenomenon the halo analysis measures directly---so the model serves as a \emph{descriptive projection}, and every severity conclusion is verified on raw cell means and bootstrap CIs; the full fit table ships with the artifact.\\[4pt]
\textbf{ASAP calibrations} are per set (partial-credit over each set's per-rater range; the two trained raters enter as rater levels); cross-set quantities are standardized before comparison.\\[4pt]
\textbf{G/D-studies.} Components are ANOVA expected-mean-squares estimators for (essay:prompt)$\times$judge with replications nested in essay$\times$judge cells (the design is balanced and complete, where these estimators are exact). Per-judge dependability $\phi(k) = \sigma^2_{\text{essay}}/(\sigma^2_{\text{essay}} + \sigma^2_{\text{rep},j}/k)$ uses each judge's own replication variance (heteroscedastic across judges by a factor of $\sim$6).\\[4pt]
\textbf{Identity canaries.} Twelve fixed essays (disjoint from all analysis samples) are scored by every judge at temperature 0 in three sweeps across the collection window (432 attempted calls, including the gated judge's refused requests, captured as error records); the per-judge response distribution is an instrument fingerprint. A canary \emph{distribution shift} triggers a halt-and-void rule. The \texttt{sonnet-4} legacy gate was an availability failure, not a distribution shift, so that rule did not apply: it was detected at the next sweep, collection continued with 11 judges, and the ENEM data were retained under D-017. Its single completed 12/12-clean sweep fell 2.8 minutes before its last ENEM call, so its window is canary-certified on the closing side only; its start was never certified.\\[4pt]
\textbf{Treatment-arm specification} (Section~\ref{sec:treat}): 200 splits; 400 held-out essays per split; anchors drawn without replacement from the complement, calibrations re-estimated per draw; monitor = paired $t$ on per-essay version deltas of $k{=}3$ rep-means at nominal $\alpha{=}.05$, false alarm calibrated on a same-version replication split (.04--.06); QWK alarm threshold set to the 95th percentile of same-version null runs at each $m$; power from 1,000--2,000 resamples; per-judge tables with split intervals ship in the artifact.\\[4pt]
\textbf{Cost.} 110{,}571 captured calls ($\approx$\$460 at on-demand prices): 84{,}373 confirmatory-grid calls, 12{,}873 pre-registered auxiliary/canary/probe/pilot calls (together the 97{,}246 pre-registered total), plus 13{,}325 calls in the three post-registration robustness/treatment cells; the artifact's manifest closes the accounting.

\section{Pre-Registration Deviations (complete)}
\label{app:dev}
Deviation identifiers come from the project decision log; operational entries (D-001--D-015, D-018, D-023--D-031: licensing, infrastructure, logistics, review rounds) are excluded, and every entry affecting an analysis or claim is listed below, none omitted. The v1.1 amendment was drafted from independent design reviews of the registration text alone; at that point the 30-essay pilot's descriptives had been seen, but no statistic of any kind had been computed on the main grids.
\begin{itemize}
\item \textbf{D-016} (pre-computation): H3 human baseline switched from unverifiable paywalled published values to the design-matched ASAP set 7--8 trained-rater statistic, frozen before any judge residual was computed.
\item \textbf{D-017}: \texttt{sonnet-4} legacy-gated by provider mid-study (canary-detected); ENEM data retained (clean window), no ASAP leg, pair not substituted; A6 reported with and without it.
\item \textbf{D-019}: G-study estimator REML$\to$ANOVA-EMS (balanced complete design; definitions unchanged).
\item \textbf{D-020}: H1 evaluated on the pre-registered bootstrap (JMLE) scale; MML agrees on ordering ($\tau{=}1.0$) on a compressed scale (both reported).
\item \textbf{D-021}: first halo implementation (raw-vs-residual mismatch) caught in QA pre-freeze; final method uses one matched statistic on both sides.
\item \textbf{D-022}: pre-writing results audit; nulls foregrounded, H3 demoted, H6 failed-as-registered, per-instrument H5b, EN paraphrase failure disclosed.
\item \textbf{D-027/D-028} (post-registration robustness cells): same-instrument trait cell (reversed H3's direction); full-panel temperature-0 anchor; both exploratory wherever used.
\item \textbf{D-032} (post-registration treatment arm): calibration bake-off, cutscore misclassification, version-boundary simulation, monitoring power (Section~\ref{sec:treat}); existing data, none confirmatory.
\item \textbf{D-033} (post-registration collection): per-dimension scoring cell (125 essays $\times$ 5 competencies $\times$ 6 cheap judges = 3,750 calls) testing the separate-calls halo remedy; exploratory.
\end{itemize}

\begingroup\renewcommand{\bibliographytypesize}{\scriptsize}
\bibliography{refs}

@book{brennan2001,
  author = {Brennan, Robert L.},
  title = {Generalizability Theory},
  publisher = {Springer},
  address = {New York},
  year = {2001}
}

@book{eckes2015,
  author = {Eckes, Thomas},
  title = {Introduction to Many-Facet {R}asch Measurement},
  edition = {2nd},
  publisher = {Peter Lang},
  address = {Frankfurt},
  year = {2015}
}

@article{engelhard1994,
  author = {Engelhard, George},
  title = {Examining rater errors in the assessment of written composition with a many-faceted {R}asch model},
  journal = {Journal of Educational Measurement},
  volume = {31},
  number = {2},
  pages = {93--112},
  year = {1994}
}

@misc{hewlett2012,
  author = {{Hewlett Foundation}},
  title = {The {A}utomated {S}tudent {A}ssessment {P}rize ({ASAP}) essay corpus},
  howpublished = {Kaggle competition dataset},
  year = {2012}
}

@book{linacre1989,
  author = {Linacre, John M.},
  title = {Many-Facet {R}asch Measurement},
  publisher = {MESA Press},
  address = {Chicago},
  year = {1989}
}

@inproceedings{marinho2021,
  author = {Marinho, Jeziel and Anchi{\^e}ta, Rafael and Moura, Raimundo},
  title = {Essay-{BR}: a {B}razilian corpus of essays},
  booktitle = {Anais do III Dataset Showcase Workshop},
  pages = {53--64},
  publisher = {Sociedade Brasileira de Computa{\c c}{\~a}o},
  year = {2021}
}

@article{myford2003,
  author = {Myford, Carol M. and Wolfe, Edward W.},
  title = {Detecting and measuring rater effects using many-facet {R}asch measurement: {P}art {I}},
  journal = {Journal of Applied Measurement},
  volume = {4},
  number = {4},
  pages = {386--422},
  year = {2003}
}

@article{myford2004,
  author = {Myford, Carol M. and Wolfe, Edward W.},
  title = {Detecting and measuring rater effects using many-facet {R}asch measurement: {P}art {II}},
  journal = {Journal of Applied Measurement},
  volume = {5},
  number = {2},
  pages = {189--227},
  year = {2004}
}

@article{saal1980,
  author = {Saal, Frank E. and Downey, Ronald G. and Lahey, Mary A.},
  title = {Rating the ratings: {A}ssessing the psychometric quality of rating data},
  journal = {Psychological Bulletin},
  volume = {88},
  number = {2},
  pages = {413--428},
  year = {1980}
}

@misc{gtheoryap2025,
  author = {Song, Dan and Lee, Won-Chan and Jiao, Hong},
  title = {Exploring {LLM} autoscoring reliability in large-scale writing assessments using generalizability theory},
  howpublished = {arXiv:2507.19980},
  year = {2025}
}

@misc{judgedatasheet2026,
  author = {Usami, Hiroyasu and Hara, Keisuke and Tsuboi, Ayato and Matsuda, Naohiko},
  title = {{LLM} judges have dark current: {A} psychometric datasheet for {LLM}-as-a-judge evaluation},
  howpublished = {arXiv:2606.15610},
  year = {2026}
}

@misc{irtasag2026,
  author = {Cong, Longwei and Hahn, Sonja and Gombert, Sebastian and Camus, Leon and others},
  title = {Estimating {LLM} grading ability and response difficulty in automatic short answer grading},
  howpublished = {arXiv:2605.00238},
  year = {2026}
}

@misc{agreementmetrics2026,
  author = {Rao, Delip and Callison-Burch, Chris},
  title = {Agreement metrics for {LLM}-as-judge evaluation: {W}hat to report and why},
  howpublished = {arXiv:2606.00093},
  year = {2026}
}

@misc{humanmfrm2026,
  author = {Casabianca, Jodi M. and Beiting-Parrish, Maggie},
  title = {Correcting human labels for rater effects in {AI} evaluation: {A}n item response theory approach},
  howpublished = {arXiv:2602.22585},
  year = {2026}
}

@misc{brandvar2026,
  author = {{\.Z}atuchin, Dmitrij},
  title = {Where does the noise come from? {A} variance-components decomposition of non-determinism in {LLM} brand answers},
  howpublished = {arXiv:2607.13304},
  year = {2026}
}

@misc{selfref2026,
  author = {Bann{\`o}, Stefano and Knill, Kate and Gales, Mark},
  title = {Towards self-referential analytic assessment: {A} profile-based approach to {L2} writing evaluation with {LLM}s},
  howpublished = {arXiv:2605.04298},
  year = {2026}
}

@misc{llmnothuman2026,
  author = {Mathew, Jerin George and Taher, Sumayya and Kundu, Anindita and Barbosa, Denilson},
  title = {{LLM}s do not grade essays like humans},
  howpublished = {arXiv:2603.23714},
  year = {2026}
}

@book{shermis2013,
  author = {Shermis, Mark D. and Burstein, Jill},
  title = {Handbook of Automated Essay Evaluation: Current Applications and New Directions},
  publisher = {Routledge},
  address = {New York},
  year = {2013}
}

@article{attali2006,
  author = {Attali, Yigal and Burstein, Jill},
  title = {Automated essay scoring with e-rater v.2},
  journal = {Journal of Technology, Learning, and Assessment},
  volume = {4},
  number = {3},
  year = {2006}
}

@inproceedings{gibson2017,
  author = {Gibson, Andrew and Aitken, Adam and S{\'a}ndor, {\'A}gnes and Buckingham Shum, Simon and Tsingos-Lucas, Cherie and Knight, Simon},
  title = {Reflective writing analytics for actionable feedback},
  booktitle = {Proceedings of the Seventh International Conference on Learning Analytics \& Knowledge ({LAK} '17)},
  pages = {153--162},
  publisher = {ACM},
  year = {2017}
}

@article{knight2020,
  author = {Knight, Simon and Shibani, Antonette and Abel, Sophie and Gibson, Andrew and Ryan, Philippa and others},
  title = {{AcaWriter}: A learning analytics tool for formative feedback on academic writing},
  journal = {Journal of Writing Research},
  volume = {12},
  number = {1},
  pages = {141--186},
  year = {2020}
}

@article{ramesh2022,
  author = {Ramesh, Dadi and Sanampudi, Suresh Kumar},
  title = {An automated essay scoring systems: a systematic literature review},
  journal = {Artificial Intelligence Review},
  volume = {55},
  pages = {2495--2527},
  year = {2022}
}
\endgroup

\end{document}